\pdfoutput=1
\documentclass[11pt]{article}

\usepackage[preprint]{acl}

\usepackage{times}
\usepackage{latexsym}
\usepackage[T1]{fontenc}
\usepackage[utf8]{inputenc}
\usepackage{microtype}
\usepackage{inconsolata}
\usepackage{graphicx}
\usepackage{amsmath}
\usepackage{amssymb}
\usepackage{booktabs}

\title{Planting a Latent Variable in Natural-Looking Text: a More Realistic Test of Belief States in LLMs and Their Link to Concept Geometry}

\author{Alexandru-Iulius Jerpelea \\
  Columbia University \\
  \texttt{aij2115@columbia.edu}}

\begin{document}
\maketitle

\begin{abstract}
LLMs are thought to track ``belief states,'' i.e., running probability distributions over the latent variables that govern language \citep{shai2024belief, sarfati2026shape}, but so far this has only been comprehensively demonstrated on toy synthetic data and in a few isolated case studies. It has also never been empirically connected to the geometry of LLM features (the concepts interpretability finds in model activations). In this work, we plant a controllable latent variable inside natural-looking text. An LLM teacher writes ordinary text while we ``subliminally'' steer\footnote{Term borrowed from \citet{morgulis2026subliminal}.} it along one of $K = 8$ unrelated sparse autoencoder directions at each token, with the active directions following a ring-shaped Markov chain. A small transformer model trained on this corpus does indeed track the Bayesian posterior belief about our planted latent variable. Moreover, it also arranges the 8 states themselves on a ring, in the exact order of the Markov chain, which is supporting evidence that a concept's geometry can be formed by the statistical dynamics of the latent variable behind it.
\end{abstract}

\section{Introduction}

It is thought that natural language is governed by a set of latent variables, so a good predictor of language must infer their state, gaining a sort of world model.\footnote{Ilya Sutskever, e.g., in a podcast: \url{https://www.youtube.com/watch?v=YEUclZdj_Sc}} \citet{shai2024belief} train small transformers on token sequences emitted by a Hidden Markov Model (HMM), and find that the residual stream linearly encodes a belief state, i.e., a running probability distribution over the state of the HMM, updating at each token.

But the paper's setup is purely synthetic, as the HMM dictates the whole toy language, and just because transformers can model belief states, it does not mean they will in a real natural language setting.\footnote{Notably, related work like \citet{sarfati2026shape} shows that a pretrained LLM fed a sequence of numbers drawn from a Gaussian will track its distribution's mean and variance, and adapt when they change. But this is just a singular case study on in-context inference, with a considerably different setup, so we let it aside as a proof to \citeauthor{shai2024belief}'s hypothesis.} A realistic controlled experiment is hard to set up as we can't model English with a bunch of Markov Models. LLMs would not be needed if we could.

\begin{figure}[t]
  \centering
  \includegraphics[width=0.8\columnwidth]{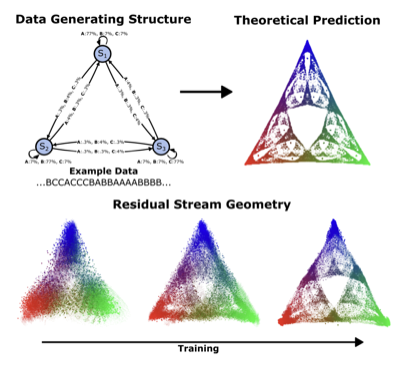}
  \caption{Figure from \citet{shai2024belief}, showing how given a data-generating process, a model learns to model optimal beliefs sitting on the probability simplex at any token.}
  \label{fig:shai}
\end{figure}

In parallel, the interpretability literature is documenting the geometry of concepts. The Linear Representation Hypothesis \citep{park2023linear}, which proposes features as linear directions, has been weakened to allow concepts to sit on low-dimensional manifolds, like days of the week on a circle, or calendar years on a helix \citep{engels2024not}. There is no full proof of why these concept manifolds form.\footnote{There are some attempts at explaining manifold geometry, like \citet{modell2025origins} tying distance and relatedness in representation space, or \citet{karkada2026symmetry} finding a correlation between concept geometry and training corpus statistics.}

In a follow-up paper, \citet{shai2026factored} show that when text is modeled by multiple HMMs at once (not just a singular latent variable as before, but multiple at once), the transformer learns a belief state per factor, each living in its own near-orthogonal subspace. The authors suggest that multi-dimensional feature manifolds could be stemming from here. However, this requires further explanation. \emph{\textbf{First of all}, their models are once again trained on synthetic HMM tokens, which is not a sufficient argument. \textbf{Second of all}, probing the belief state, which gives a probability distribution over a latent variable's values, is not the same as asking how the values themselves that the latent variable takes are geometrically arranged.}

\begin{figure}[t]
  \centering
  \includegraphics[width=\columnwidth]{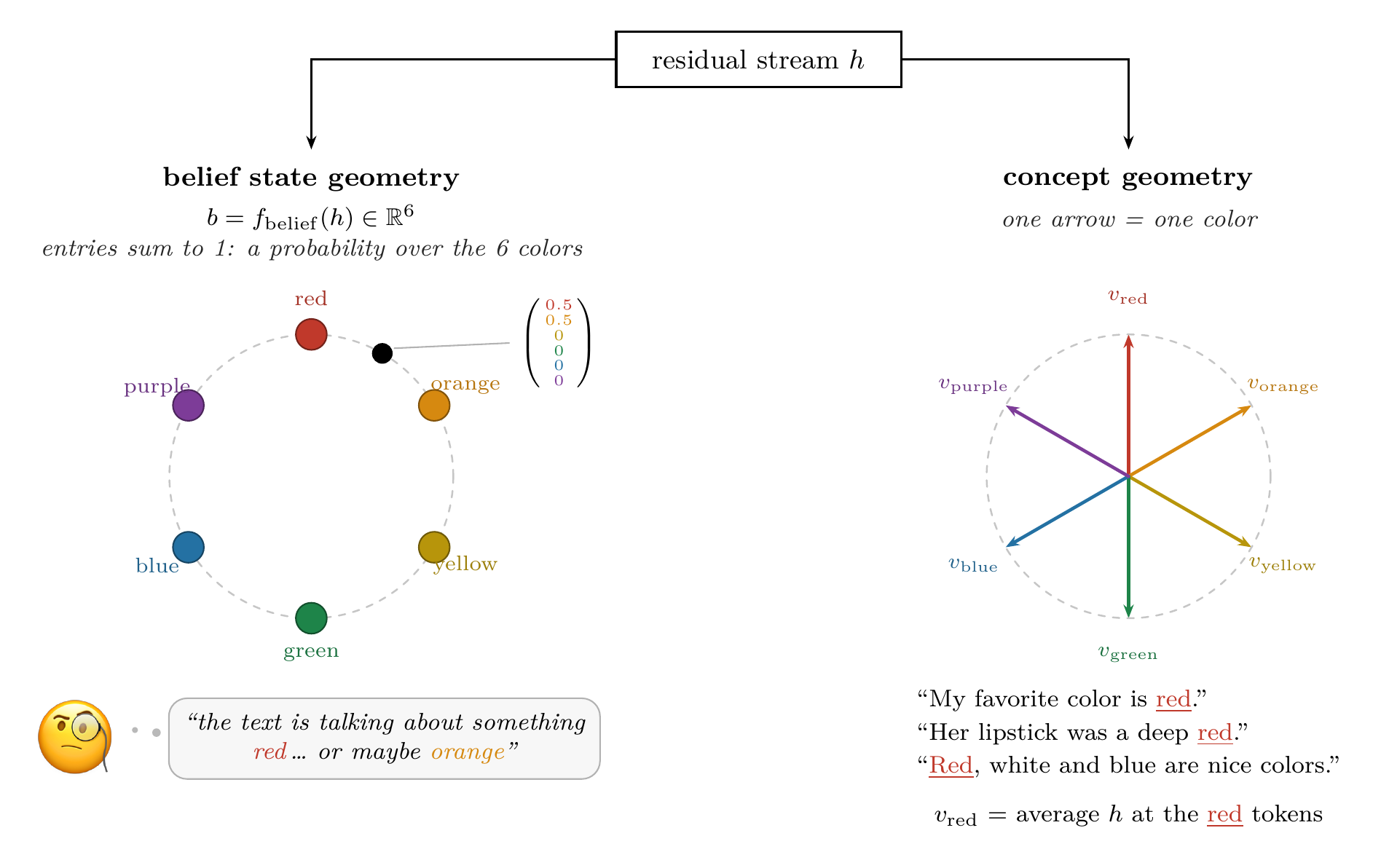}
  \caption{Belief state geometry (each point a probability vector over the latent variable's values) versus concept geometry (the arrangement of the values themselves), both read from the residual stream. The two are different objects.}
  \label{fig:beliefvsconcept}
\end{figure}

Figure~\ref{fig:beliefvsconcept} better illustrates the difference between belief states and concepts/features within the residual stream. It is unclear whether belief dynamics are actually correlated with concept geometry.

This paper tries to tie both ends by introducing a method for generating natural-looking text influenced by a latent variable with a Markov structure of our choosing. We use an LLM teacher that writes ordinary text while we ``subliminally'' steer it \citep{morgulis2026subliminal} along one of $K = 8$ unrelated and orthogonal sparse autoencoder directions. Switching from one autoencoder direction to another is dictated by a ring-shaped Markov chain that we fully control. We then train a small transformer from scratch on the generated corpus and ask two questions. Does it track the belief state of our variable? (Yes, confirming the \citet{shai2026factored} hypothesis in a more realistic scenario). And do the 8 states themselves inherit the geometry of the Markov Model? (Yes, they sit on a ring, in the exact neighbor order).

\section{Method}

The main difficulty in generating natural-looking text governed by controllable latent variables is that it's very hard to write down the latent variables of natural language as a set of HMMs.\footnote{Note that any tokenized data generating process can be theoretically modeled by an HMM (given the right conditions). Thus, it is not unreasonable to think of language being governed by an HMM, or a set of HMMs.} If we could do that, we would not need LLMs. But that's exactly the method's intuition! LLMs already model language with high accuracy, and although they are not fully explainable or controllable, they are steerable to some degree. And for our goal, one controlled latent variable is enough. So instead of trying to write down language's latent variables, we use steering to add one of our own: at each token, we steer the teacher along one of $K = 8$ uncorrelated and orthogonal SAE directions, while generating otherwise ordinary text. We only steer along one SAE direction at a time, given by a Markov chain which updates at every token. We then train a student model from scratch on the generated corpus.

One could consider the SAE directions to be 8 separate \emph{latent variables} of natural language. However, by imposing this artificial dynamic over them, we create a \emph{new} latent variable, and the 8 directions become \emph{values} that the \emph{new} latent variable can take. We also specifically pick the 8 directions to be orthogonal and uncorrelated, such that our structure could not have come from anywhere else.

Note that steering is known to inject durable, learnable structure into generated text, even if the text is seemingly unrelated to the steering direction (which is what we aim for). In their study, \citet{morgulis2026subliminal} call this mechanism ``subliminal steering,'' finding that fine-tuning a student on random number sequences written by a steered teacher makes the student inherit the teacher's steering direction.

\begin{figure*}[t]
  \centering
  \includegraphics[width=0.92\textwidth]{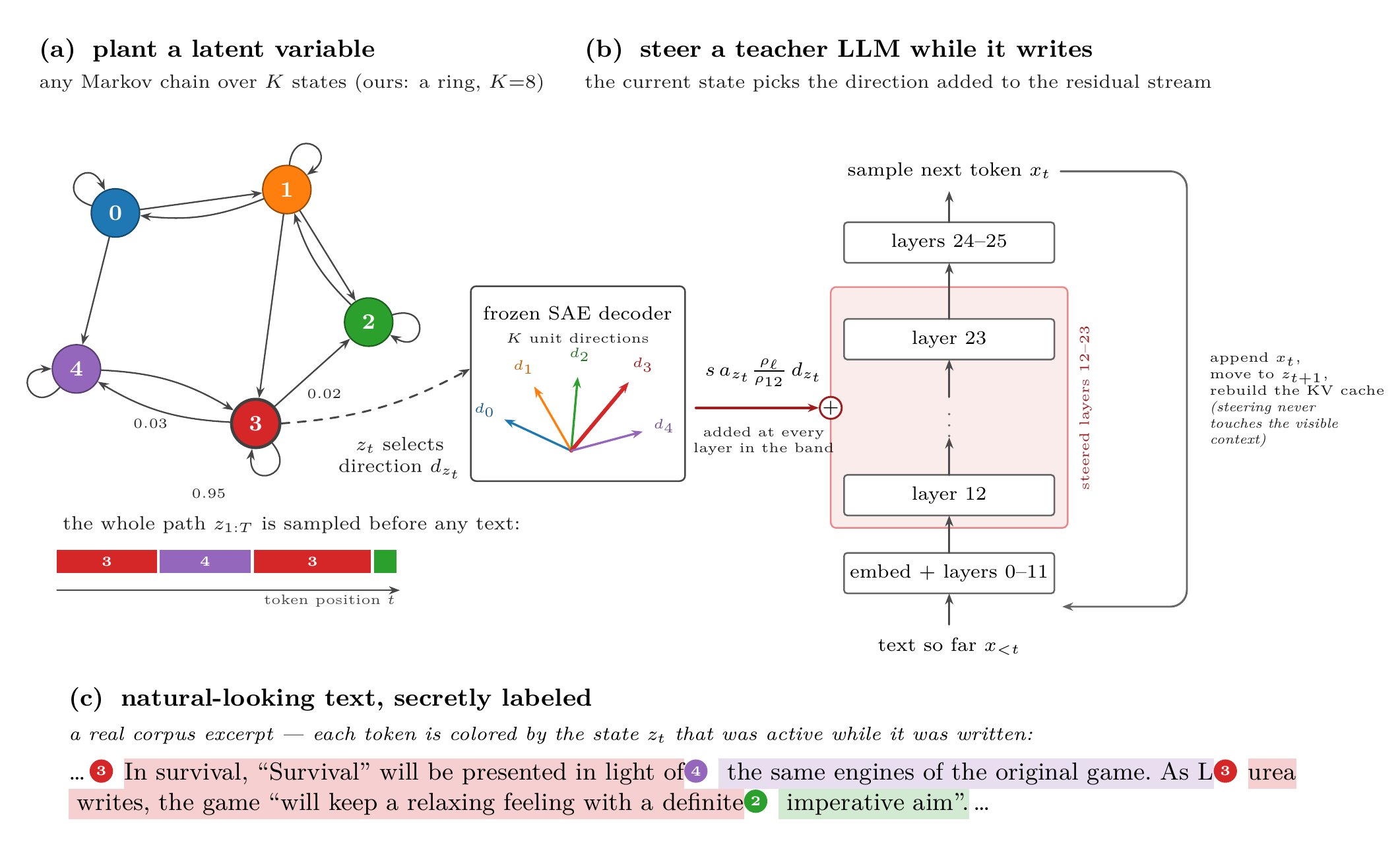}
  \caption{Method summary: (a) we plant a latent variable as a Markov chain over $K$ states; (b) we steer a teacher LLM while it writes, by adding the current state's SAE decoder direction to the residual stream; (c) the result is natural-looking text, secretly labeled per token.}
  \label{fig:method}
\end{figure*}

Figure~\ref{fig:method} gives a visual summary of our proposed algorithm.

\subsection{The Generative Process}

The teacher model is a Gemma-2-2B (base) \citep{gemmateam2024gemma2}, which we couple with a pretrained sparse autoencoder (Gemma Scope, 16K latents; \citealp{lieberum2024gemmascope}) for the residual stream of a middle layer (layer \#12). From these we select $K = 8$ uncorrelated latents (the filtering is described below). Each selected latent $i$ has a unit decoder direction $d_i$ and an average firing magnitude $a_i$. Steering along latent $i$ means adding $s \cdot a_i \cdot \frac{\rho_\ell}{\rho_{12}} \cdot d_i$ to the residual stream at layers 12 through 23 $(L-2)$,\footnote{Admittedly, we are using an SAE direction computed at layer 12 to steer the residual stream at other layers, too. We have done this after an experiment which revealed that artificial steering at just one level allows the model to undo our action at later levels. One might raise the problem of the SAE direction being out-of-distribution for upper layers, but that's fine considering that the residual stream is a sum of activations at all layers: one can imagine that at upper layers we are just nudging the piece corresponding to layer 12.} where $\rho_\ell$ is the average residual-stream norm at layer $\ell$, so compensates for the stream's growing norm. Lastly, $s = 1.5$ was chosen by a sweep, such that the injection is strong enough, yet does not degenerate the quality of the text.

\begin{figure*}[t]
  \centering
  \includegraphics[width=0.85\textwidth]{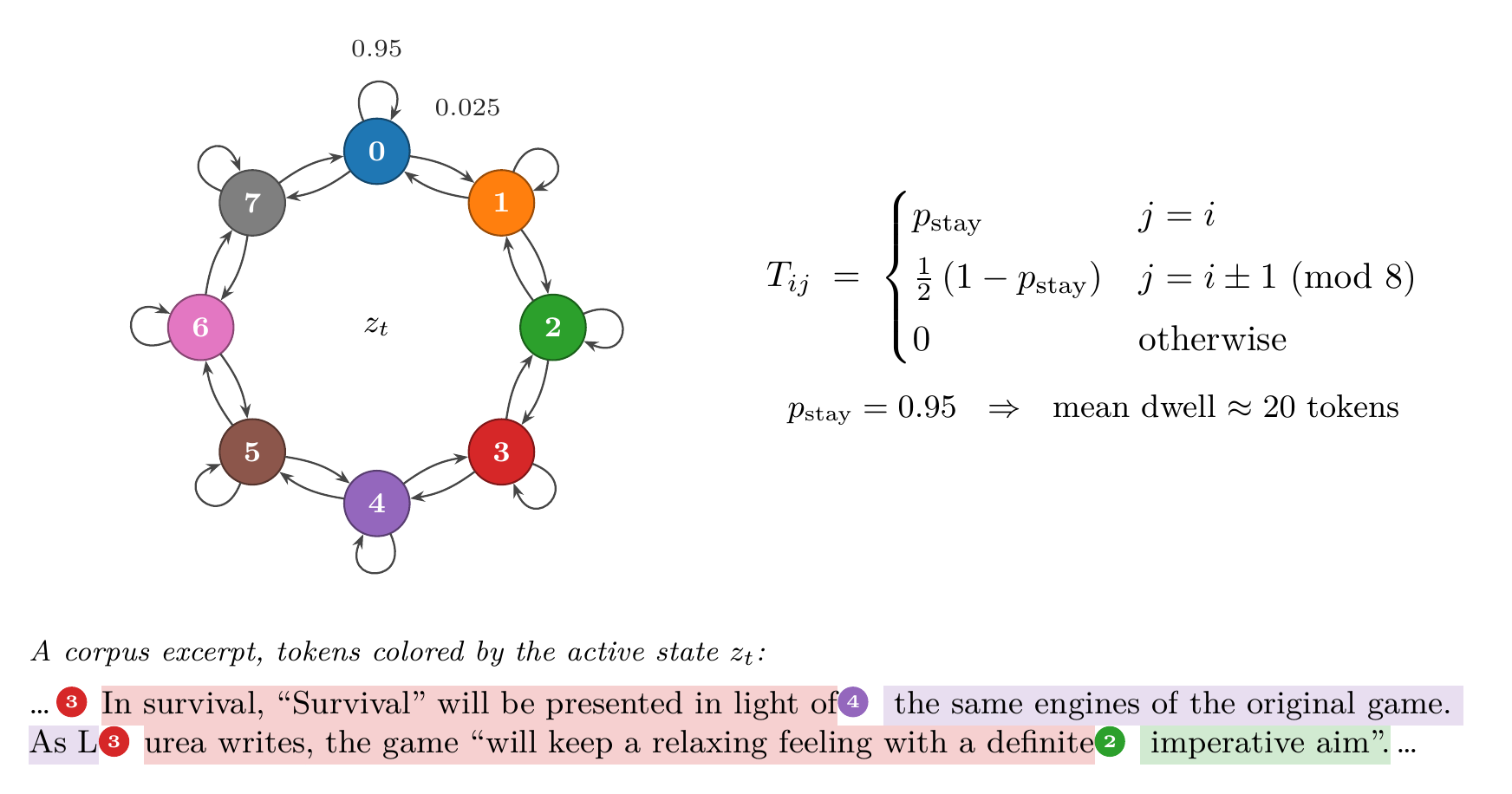}
  \caption{The 8-state ring Markov chain with stay probability $0.95$ and its transition matrix, plus a corpus excerpt with tokens colored by the active state.}
  \label{fig:ring}
\end{figure*}

The corpus consists of independent \emph{documents} of 256 tokens each. The first few tokens come from a pre-selected set of ``openers,'' and the teacher writes the rest. At every token, we steer the teacher along exactly one of the 8 directions. We write $z_t \in \{1, \ldots, 8\}$ for the direction active while the token at position $t$ is sampled. The 8 directions, together with the dynamics we impose on them, constitute the synthetic latent variable. One can think of it as a hidden ``concept'' governing the text, like the emotion variable drifting through a story (see \citealp{bigelow2026stories}), except that here we control the variable and hold its ground truth, instead of finding it in the wild. The 8 directions themselves are not new, since each is a regular feature the teacher already has (intentionally unrelated and orthogonal to the others). The latent variable they form together is new though, as its possible states are only tied to their neighbors by our transition matrix. The latent variable takes a walk along the ring shown in Figure~\ref{fig:ring}.

At $0.95$ ``stay'' probability, the variable dwells $\sim$20 tokens per state before hopping to the neighbor. Each document's path $z_1, \ldots, z_{256}$ is sampled before generation; thus, nothing else in the LM could plant the ring geometry.

Very importantly, the KV cache is always rebuilt at each token. Steering never touches the cached context, so that each token only depends on the current visible text and the latent variable. This will be very important in order to compute the outputs of an ideal Bayesian inference model, to which we will compare our student model's belief state.

\subsection{The Corpus}

We generate 400,000 documents $\times$ 256 tokens (the first 12 tokens are taken from a pre-selected set of ``openers''), approx. $\sim$102M tokens. We also generate a control corpus with steering off, and we train a control student on it. Analysis will run on both students.

\subsection{Selecting the SAE Latents}

Out of the 16K SAE latents, we drop the ones that are dead, fire too often, or mostly fire on punctuation and position. We don't really care about what our chosen latents mean, and aim for them to be as hard to eyeball as possible (so that they are more ``latent''). We also make sure that in our selection, the latents are near-orthogonal, don't co-activate on natural text, and are causal (i.e., steering along them actually modifies the token distribution).

\subsection{The Optimal Observer}

At each token position in a document, we measure how well the state of the latent variable can be tracked by an observer who knows everything, i.e., knows the transition matrix of the 8 latents, and the softmax token distribution of the steered teacher model given any text.

Note that we can token label our transition matrix just like \citet{shai2024belief}'s formalisms, so we truly have an HMM:
\begin{equation}
T^{(x)}_{ij} = T_{ij} \cdot \mathrm{Teacher}_j(x \mid x_{<t}),
\end{equation}
where $\mathit{Teacher}_j$ is just the LLM with $z_t = j$.

We call our proposed reader the \textbf{optimal observer}, which performs Bayes inference over the operator defined above, maintaining the exact posterior $b_t(i) = P(z_t = i \mid x_{1:t})$ at every token. Since $b_t$ is a probability distribution over the 8 states, it is a point in the probability simplex $\Delta^7$. The posteriors get updated as such at each token position $t$:
\begin{equation}
b_t(j) \propto \textstyle\sum_i b_{t-1}(i) \cdot T^{(x_t)}_{ij}
\end{equation}

Because the KV cache is reset at every token, the next token only depends on the visible text and the current state of the model, thus making the optimal observer computable at every point. We use $b_t$ at every token as the probe target for our student.

\begin{figure}[t]
  \centering
  \includegraphics[width=0.75\columnwidth]{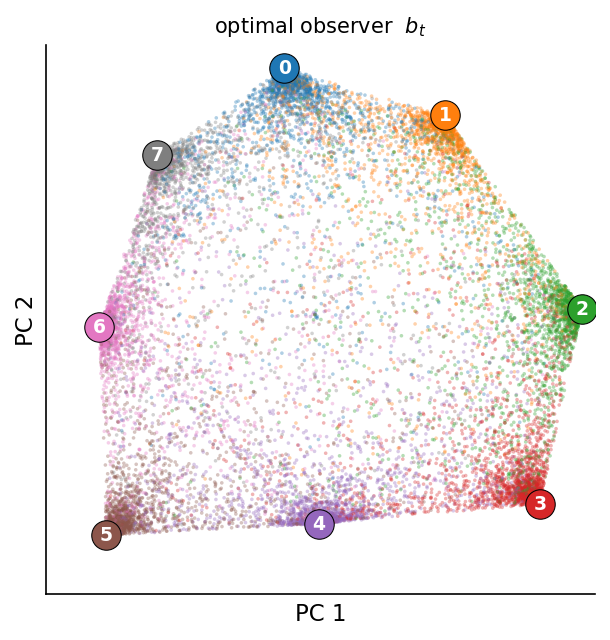}
  \caption{PCA of the optimal observer's posteriors: the point cloud fills a ring-shaped region with the 8 states in ring order at the rim.}
  \label{fig:obspca}
\end{figure}

The optimal belief $b_t$ sits on a 7-dimensional probability simplex. But if we PCA on it we find that the two principal components reflect the ring structure of $T$ (Figure~\ref{fig:obspca}).

\subsection{The Student}

We use a 110M-parameter Llama-style transformer (12 layers, width 768), trained from scratch on the corpus tokens and nothing else ($\sim$5 epochs, 32k vocab). Unlike the optimal observer, the student has no access to the teacher model, so $z$ reaches it intertwined with all the other latent variables of the Gemma model. We ask whether the factorized belief state over $z$ emerges anyway, and we also explore the geometry of the latent variable.

\begin{figure*}[t]
  \centering
  \includegraphics[width=0.85\textwidth]{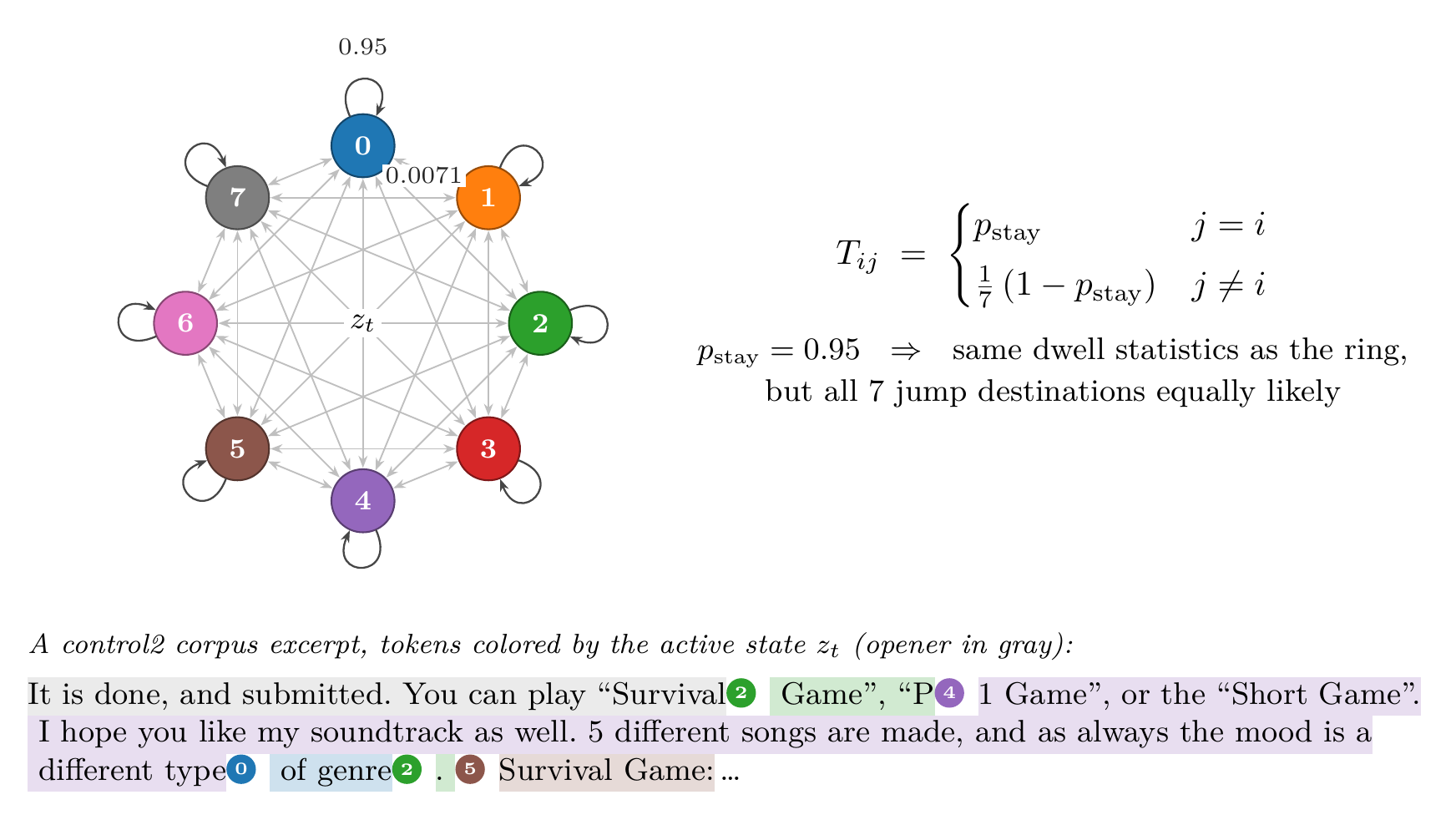}
  \caption{The \emph{control2} chain: same stay probability, but uniform jumps to any of the other 7 states, plus a \emph{control2} corpus excerpt with tokens colored by the active state.}
  \label{fig:control2}
\end{figure*}

We also train two control students. \emph{control1} was trained on a completely unsteered corpus from the same teacher. \emph{control2} was trained on a corpus steered by the same 8 latents with the same $p_{stay}$, but on a state switch, the latent variable jumps uniformly to any of the other 7 states instead of to neighbors on the ring (Figure~\ref{fig:control2}). It has the same exposure to the 8 features and incentive to track them, but no transition map to learn.

\section{Results}

Recall that this work has two objectives: 1.\ to confirm that transformers learn belief states over latent variables in a more realistic setting than the toy setup of \citet{shai2026factored}, and 2.\ to find the relationship between the geometry of a latent variable itself (\emph{which is different than the geometry of the belief state!}) and its underlying statistical dynamics.

\subsection{The Student Tracks the Belief States}

Let $h^{(\ell)}_t \in \mathbb{R}^{d_{model}}$ be the student's residual stream at layer $\ell$ and token position $t$. We want to test whether the belief state can be read out linearly out of this vector, so we fit a \emph{ridge regression} from the hidden states to the posteriors of the optimal observer $b_t \in \Delta^7$, at each token: $\hat{b}_t = W h^{(\ell)}_t + c$.

\begin{figure*}[t]
  \centering
  \includegraphics[width=0.95\textwidth]{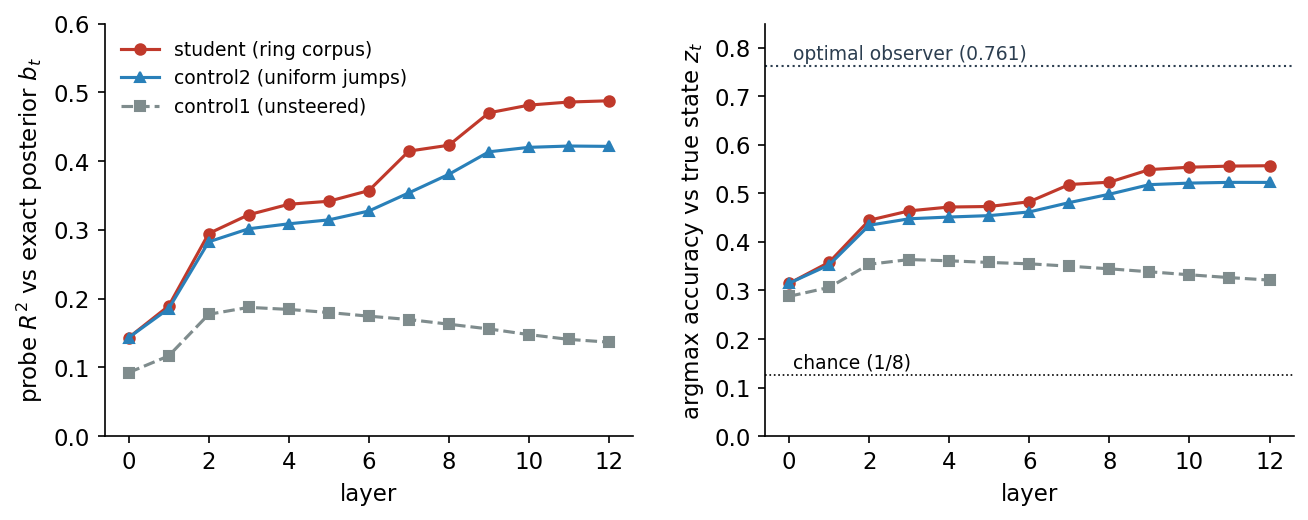}
  \caption{Per-layer probe $R^2$ and argmax accuracy for the ring student, \emph{control2}, and \emph{control1}, with the optimal observer ceiling at $0.762$.}
  \label{fig:probe}
\end{figure*}

We fit the map on 4,000 held-out documents from our circularly steered corpus, and evaluate it on 1,000 more. Per layer, we report the regression's $R^2$ against $b_t$, and the accuracy of $\arg\max_i \hat{b}_t(i)$ against the true state $z_t$, averaged over all token positions (Figure~\ref{fig:probe}).

Indeed, the belief posterior does live in the residual stream. The best single layer reaches $R^2 = 0.49$, with argmax accuracy of the actual state of $0.577$, which is approximately three quarters of the optimal observer's ceiling of $0.762$. This is significant given that the student had no direct access to the steered teacher model, as the optimal observer did.

As expected, \emph{control1} does not map the belief state (low $R^2$), but does guess the current state above chance. The SAE directions we steer along are natural in language, so ordinary text has coincidentally exposed it to them without even steering. But note that \emph{control2} reaches a significant $R^2 = 0.41$ on the same documents, also with high argmax accuracy ($0.52$). This is also expected, as \emph{control2} has seen the eight states just as much, so after enough tokens in one state, it can decently estimate which state is active. In other words, after enough tokens in a state, $b_t$ converges towards a one-hot vector, so there is nothing intrinsic about the ring geometry in it.

\begin{figure*}[t]
  \centering
  \includegraphics[width=0.95\textwidth]{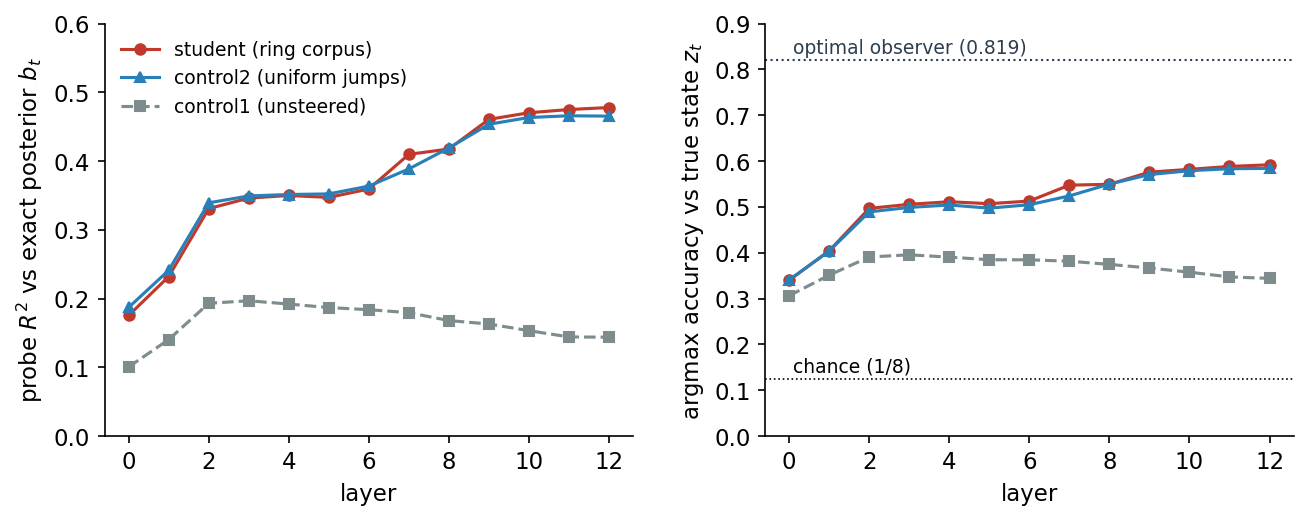}
  \caption{Per-layer probe $R^2$ and argmax accuracy restricted to first-dwell tokens: the ring student and \emph{control2} curves coincide.}
  \label{fig:firstdwell}
\end{figure*}

We believe the $R^2$ ($\approx 0.08$) difference between the two is the advantage that the student model has from knowing how the previous state constrains the next one, i.e., internalizing the HMM's structure. To confirm this, we re-evaluate on the \emph{first dwell} only, where we define the \emph{first dwell} of a document to be the prefix where the latent variable has not yet switched from its initial state. As shown in Figure~\ref{fig:firstdwell}, in this scenario, \emph{control2} and our student become indistinguishable. As there is no prior information, the transition map is useless even to the optimal observer, so this ablation is a positive result towards showing that the student calculates belief states.

\begin{figure}[t]
  \centering
  \includegraphics[width=\columnwidth]{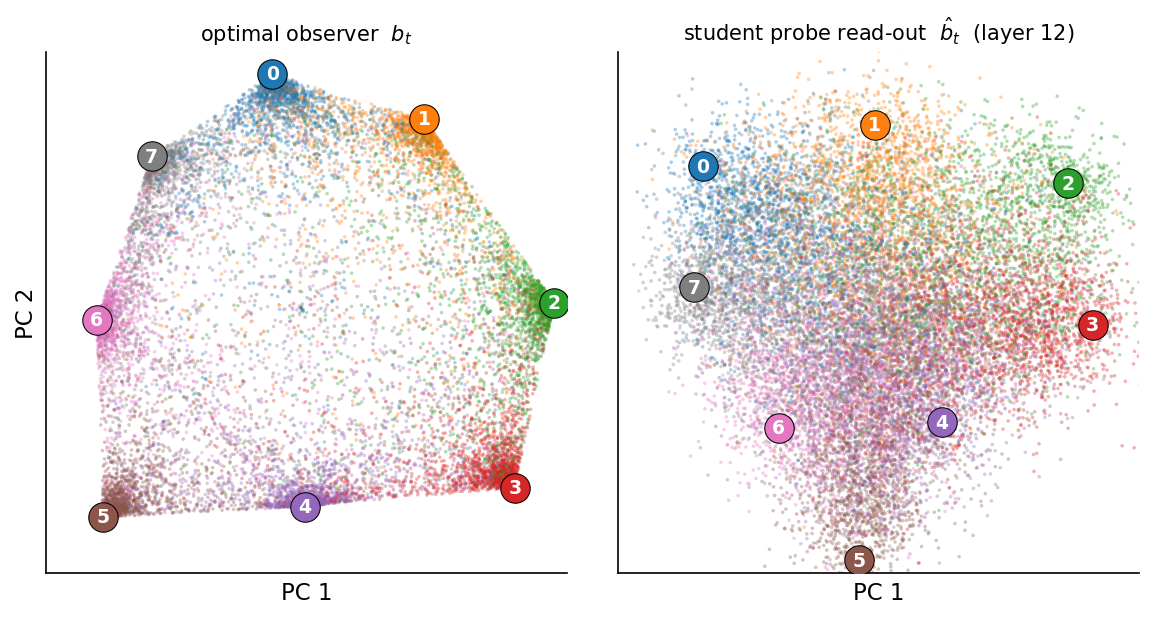}
  \caption{Side-by-side PCA of the optimal observer's posteriors and the student's probe read-out at layer 12: same ring arrangement, noisier cloud for the student.}
  \label{fig:studentpca}
\end{figure}

For our student, we can also look at the probe's read-out $\hat{b}_t$ in the same PCA view we used for the optimal observer. The student's belief state is a noisier version of the optimal observer's (Figure~\ref{fig:studentpca}).

\subsection{The Eight States of the Latent Variable Sit on a Ring}

So far we have shown that the student does hold beliefs resembling $b_t$. But \emph{control2} also does a decent job at inferring the belief state, because in our setup, each token carries some evidence about the current state, and, after enough tokens, that evidence alone suffices for identifying the state. Knowing the previous state only helps our student model early on, right after a state switch.

We now show where the models truly differ, which is how they represent the latent variable. Note that the geometry of the latent variable is a different object than the belief geometry above. We are no longer curious about the shape of the probability distribution, but about the shape of the states themselves. Specifically, we take the mean of the residual-stream activations for tokens in each of the 8 states (call them \emph{centroids}) and we test what shape the 8 points form. If the geometry of the dynamics is inherited, they should sit on a ring.

\begin{figure}[t]
  \centering
  \includegraphics[width=\columnwidth]{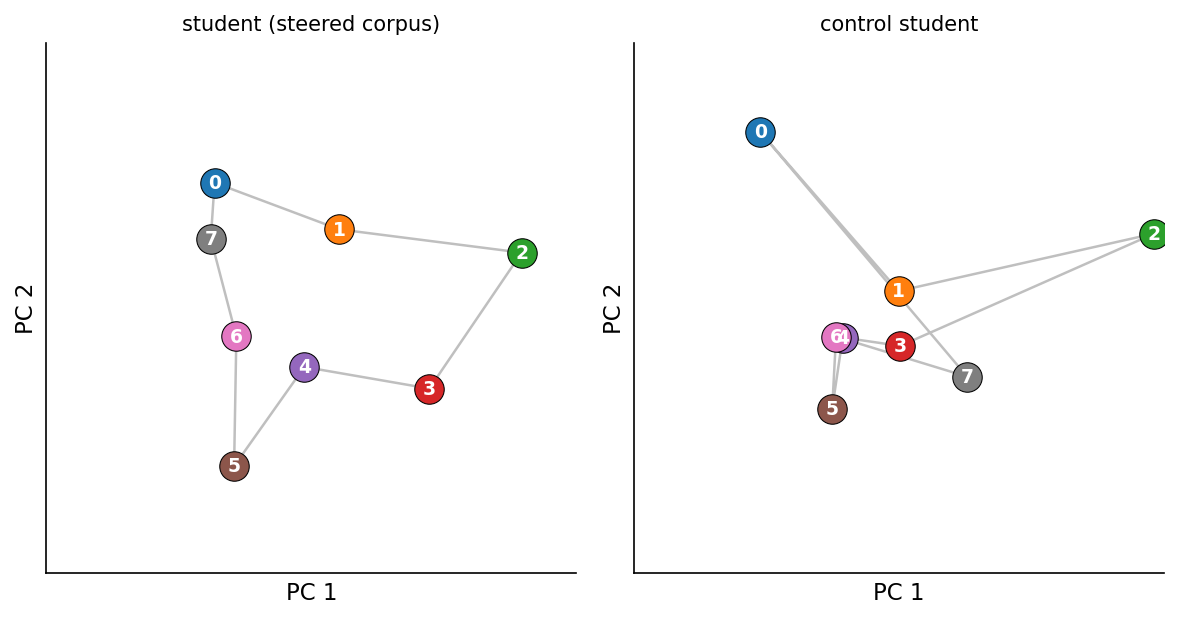}
  \caption{Top-2 PCA of the 8 state centroids: the ring student's centroids trace the ring in exact neighbor order; the control student's do not.}
  \label{fig:centroids}
\end{figure}

We project the centroids onto their top-2 principal components. In layers 9--11, the 8 centroids sit in the exact neighbor order of our Markov chain, and pairwise distance between centroids correlates with distances along the ring at $0.45$. The \emph{control1} student shows nothing (with correlation $0.11$); see Figure~\ref{fig:centroids}.

\begin{figure}[t]
  \centering
  \includegraphics[width=\columnwidth]{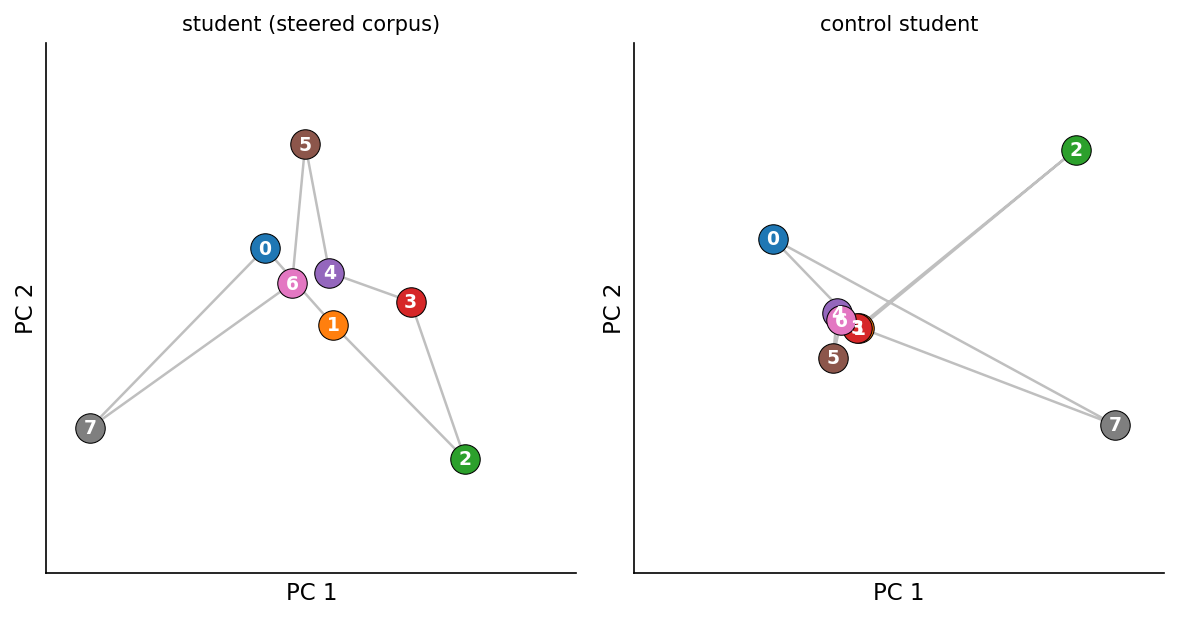}
  \caption{Top-2 PCA of first-dwell centroids: the clean ring is no longer visible in the leading principal components.}
  \label{fig:firstdwellpca}
\end{figure}

But note that this is a trap! Right after a switch from state 2 to state 3, the activations might still remember state 2 for a few tokens. So when we average all of state 3's tokens to get state 3's centroid, we are also getting some of state 2 (and state 4), because the previous state in the text is always a ring neighbor. This means that each of the 8 centroids gets dragged towards its neighbors, and that alone draws a ring. So the circle above could just be short-term memory of the context, without the model actually arranging the states on a ring. From now on, we only compute centroids on each document's \emph{first dwell} (i.e., before the latent variable ever switches states), so that no previous states leak in. In this setting, the PCA circle does not survive (Figure~\ref{fig:firstdwellpca}).

\begin{figure*}[t]
  \centering
  \includegraphics[width=0.92\textwidth]{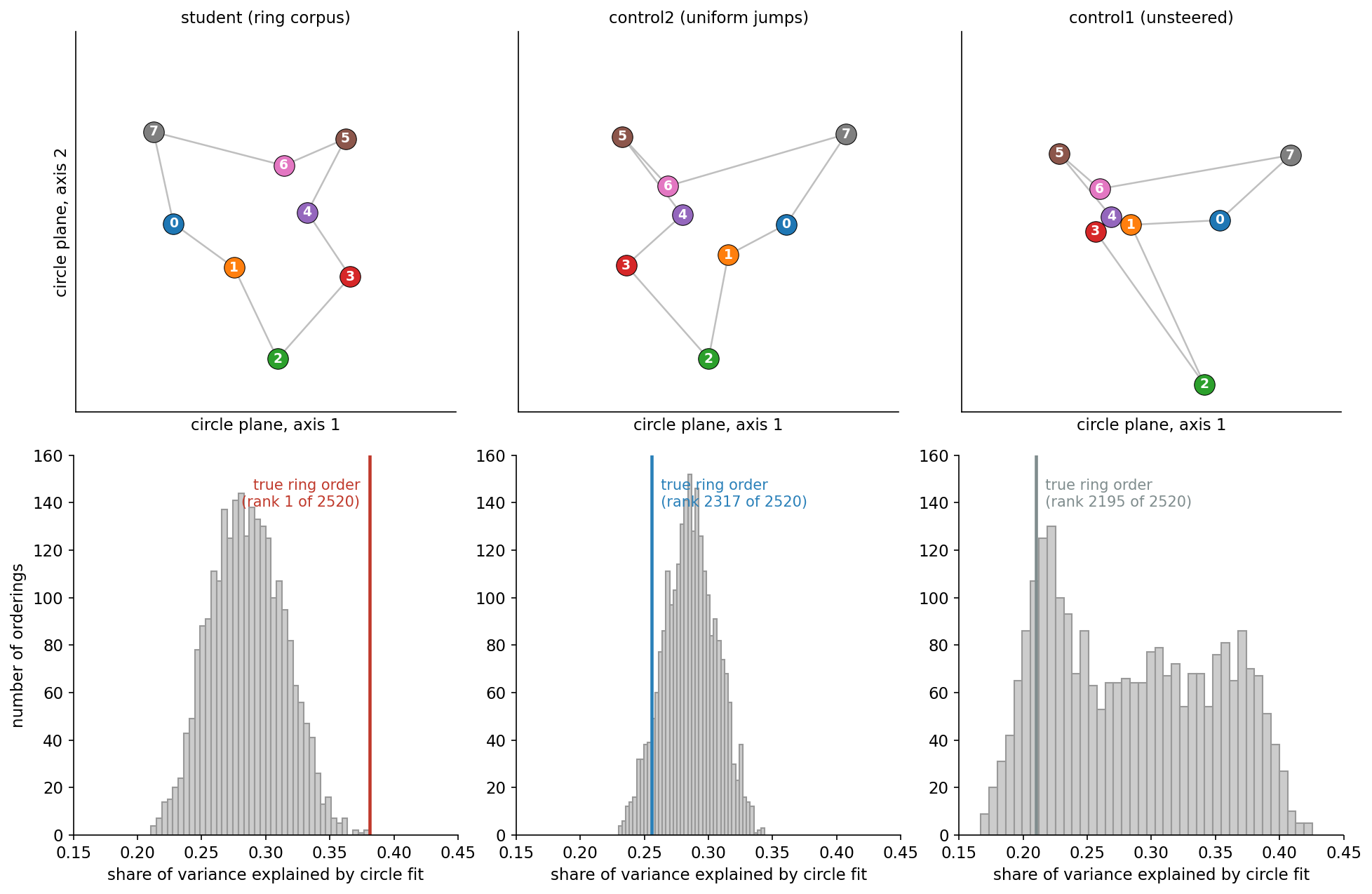}
  \caption{Circle-plane projections of first-dwell centroids for the three students, and the circle-fit permutation test over all 2520 ring orderings: the true ring ranks 1st of 2520 for the ring student only.}
  \label{fig:circlefit}
\end{figure*}

However, this does not mean the ring is gone, but rather that the ring geometry is not the principal geometry (we only looked at the first 2 PCA axes). We use Fourier analysis to directly query if there is any plane on which the centroids trace a circle, in the exact ring order of the HMM. Specifically, we look for two directions $u$ and $v$ such that:
\begin{equation}
\mu_i \approx u \cos(2\pi i/8) + v \sin(2\pi i/8),
\end{equation}
where $\mu_i$, with $i \in \{1, 2, \ldots, 8\}$, is the centroid of the $i$-th state. The fitted probes $(u, v)$ explain 38\% of the centroid variance at layer 11. For a significance test, we refit the same pattern under every possible way of arranging the 8 states on a ring (2520 orderings), and the true ring ranks 1st out of all 2520 in terms of explained variance, for our student. However, for the \emph{control1} and \emph{control2} models, the true order lands mid-distribution. So, the circle geometry is real, respecting the exact ring-order, and can be attributed to the HMM structure. Moreover, we have also finally differentiated the student from \emph{control2} (Figure~\ref{fig:circlefit}).

We think the ring geometry is not dominant simply because our latent variable is not that relevant. Most of the residual stream is spent on ordinary language features, and our variable only injects $\sim$0.05--0.1 nats of evidence per token. Moreover, the Gemma teacher surely models other latent variables too, and our 8 states could also be attributes that other latent variables land on. In other words, each centroid possibly sits in multiple geometries at once, each corresponding to other latent variables.

\begin{figure}[t]
  \centering
  \includegraphics[width=\columnwidth]{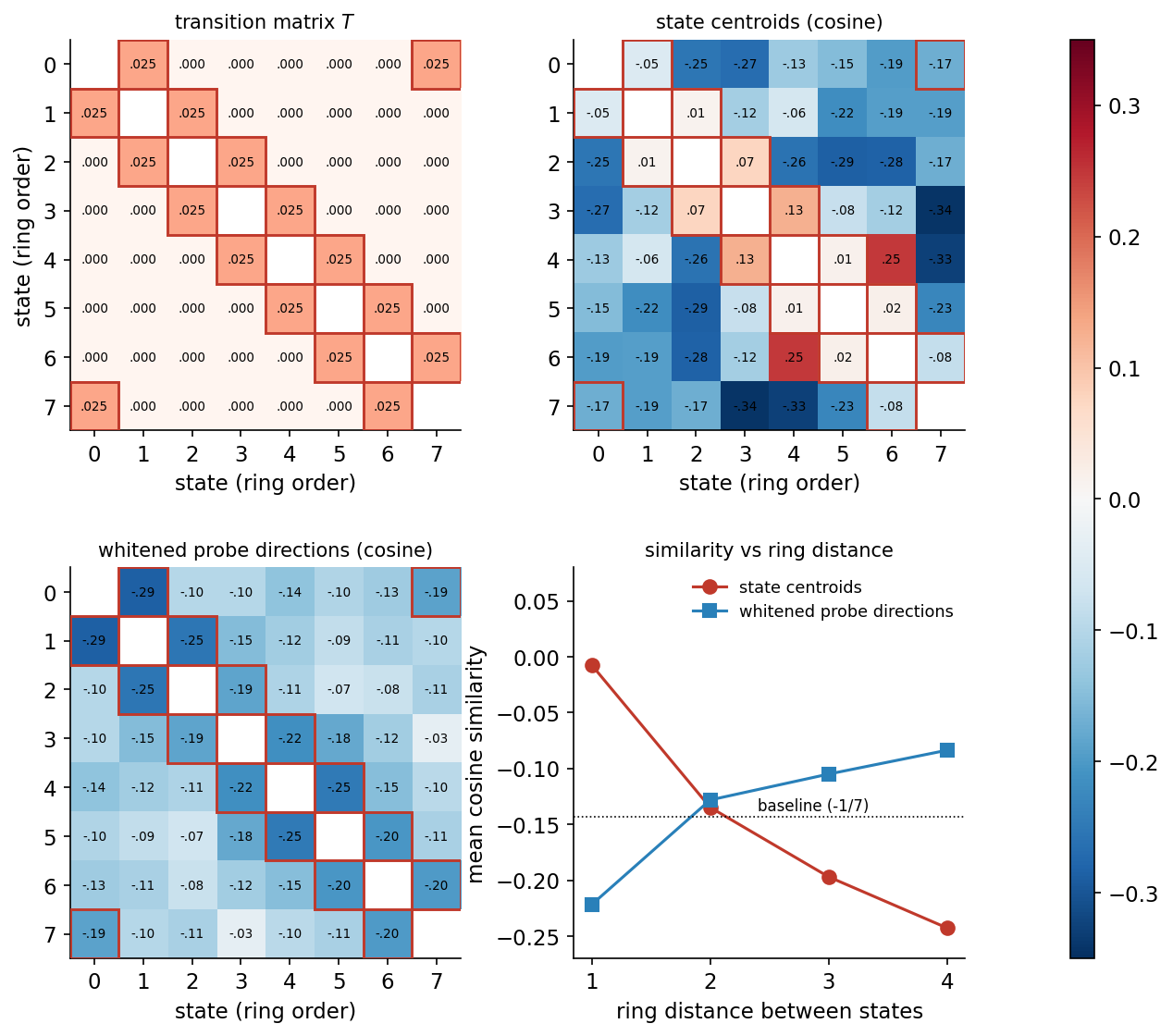}
  \caption{Heatmaps: transition matrix $T$, centroid cosine similarities warmest on the neighbor diagonals, whitened probe directions most anti-correlated there, and similarity vs.\ ring distance.}
  \label{fig:heatmaps}
\end{figure}

The ring structure is also visible in raw similarities. In Figure~\ref{fig:heatmaps}, pairwise cosine similarity between centroids is warmest on the neighbor diagonals, resembling the matrix $T$. Moreover, the columns of a whitened ridge probe are most anti-correlated on the same diagonals. This supports our case, as the probe has to spend capacity on differentiating exactly the pairs that are most similar (i.e., the ones on the ring).

\section{Conclusion}

We introduced a method for planting a latent variable inside natural-looking text, where an LLM teacher writes ordinary text while we ``subliminally'' steer it along SAE directions whose activity follows a Markov chain of our choosing.

By doing so, we confirmed that transformers learn belief state geometries in a more realistic setting than current work. Moreover, we have tied belief states to concept geometries: the geometry of a latent variable (not only the geometry of the belief state about that latent variable) is influenced by the dynamics of the data generating process.

\section{Future Work}

A first natural follow-up is to plant multiple HMMs at once with our algorithm (\citealp{shai2026factored} already do this in their toy setup), including more complicated interactions between them.

Second, many existing theories attribute feature manifolds to semantic similarity. Our setup can put the two in direct conflict: take semantically similar states (like different colors, for example) and tie them together in a Markov chain that's uncorrelated to their semantic similarity.

Third, recall that our ring was not the dominant geometry. Perhaps, each state participates in multiple other latent variables. We are curious whether a possible manifold entanglement is happening in such scenarios. We could test this directly with the multiple HMMs setup.

Fourth, we are also interested in hierarchical concepts and how they fit in this picture.

Lastly, to close the loop, we want to use this theory to make predictions about real LLMs. We want to find a latent variable (maybe something like syntax), estimate its transition dynamics from the training corpus, and predict the geometry of that concept in a real LLM, like OLMo.

\section*{Limitations}

Our setup, although more realistic than the toy setups, is still not full proof of \citeauthor{shai2024belief}'s hypothesis about belief state geometries. We impose a latent variable on language, which shows that transformers pick up such structure when it is there, and, although our setup is more natural, it suffers from the same limitation as the initial proponents of the theory.

Our evidence for the concept geometry is just correlational as we have not done any causality experiments.

We also have to expand our configurations for a more comprehensive study, i.e., we have to use more HMM structures, other choices of SAE latents, other teacher models, etc.

Lastly, we acknowledge that in order to construct a new latent variable, we had to use SAE directions, which could be latent variables themselves, or attributes that other latent variables take. This is not ideal as we were not able to fully isolate our planted latent variable (the ring geometry was not dominant), so there remains the open question of how to better inject latent variables in natural-looking text.

\bibliography{custom}

\begin{thebibliography}{11}
\providecommand{\natexlab}[1]{#1}

\bibitem[{Bigelow et~al.(2026)Bigelow, Sarfati, Wurgaft, Lewis, McGrath,
  Merullo, Geiger, and Lubana}]{bigelow2026stories}
Eric Bigelow, Rapha{\"e}l Sarfati, Daniel Wurgaft, Owen Lewis, Thomas McGrath,
  Jack Merullo, Atticus Geiger, and Ekdeep~Singh Lubana. 2026.
\newblock \href {https://arxiv.org/abs/2605.12412} {Stories in space:
  In-context learning trajectories in conceptual belief space}.
\newblock \emph{Preprint}, arXiv:2605.12412.

\bibitem[{Engels et~al.(2024)Engels, Michaud, Liao, Gurnee, and
  Tegmark}]{engels2024not}
Joshua Engels, Eric~J. Michaud, Isaac Liao, Wes Gurnee, and Max Tegmark. 2024.
\newblock \href {https://arxiv.org/abs/2405.14860} {Not all language model
  features are one-dimensionally linear}.
\newblock \emph{Preprint}, arXiv:2405.14860.

\bibitem[{{Gemma Team}(2024)}]{gemmateam2024gemma2}
{Gemma Team}. 2024.
\newblock \href {https://arxiv.org/abs/2408.00118} {Gemma 2: Improving open
  language models at a practical size}.
\newblock \emph{Preprint}, arXiv:2408.00118.

\bibitem[{Karkada et~al.(2026)Karkada, Korchinski, Nava, Wyart, and
  Bahri}]{karkada2026symmetry}
Dhruva Karkada, Daniel~J. Korchinski, Andres Nava, Matthieu Wyart, and Yasaman
  Bahri. 2026.
\newblock \href {https://arxiv.org/abs/2602.15029} {Symmetry in language
  statistics shapes the geometry of model representations}.
\newblock \emph{Preprint}, arXiv:2602.15029.

\bibitem[{Lieberum et~al.(2024)Lieberum, Rajamanoharan, Conmy, Smith, Sonnerat,
  Varma, Kram{\'a}r, Dragan, Shah, and Nanda}]{lieberum2024gemmascope}
Tom Lieberum, Senthooran Rajamanoharan, Arthur Conmy, Lewis Smith, Nicolas
  Sonnerat, Vikrant Varma, J{\'a}nos Kram{\'a}r, Anca Dragan, Rohin Shah, and
  Neel Nanda. 2024.
\newblock \href {https://arxiv.org/abs/2408.05147} {Gemma scope: Open sparse
  autoencoders everywhere all at once on gemma 2}.
\newblock \emph{Preprint}, arXiv:2408.05147.

\bibitem[{Modell et~al.(2025)Modell, Rubin-Delanchy, and
  Whiteley}]{modell2025origins}
Alexander Modell, Patrick Rubin-Delanchy, and Nick Whiteley. 2025.
\newblock \href {https://arxiv.org/abs/2505.18235} {The origins of
  representation manifolds in large language models}.
\newblock \emph{Preprint}, arXiv:2505.18235.

\bibitem[{Morgulis and Hewitt(2026)}]{morgulis2026subliminal}
George Morgulis and John Hewitt. 2026.
\newblock \href {https://arxiv.org/abs/2604.25783} {Subliminal steering:
  Stronger encoding of hidden signals}.
\newblock \emph{Preprint}, arXiv:2604.25783.

\bibitem[{Park et~al.(2023)Park, Choe, and Veitch}]{park2023linear}
Kiho Park, Yo~Joong Choe, and Victor Veitch. 2023.
\newblock \href {https://arxiv.org/abs/2311.03658} {The linear representation
  hypothesis and the geometry of large language models}.
\newblock \emph{Preprint}, arXiv:2311.03658.

\bibitem[{Sarfati et~al.(2026)Sarfati, Bigelow, Wurgaft, Boppana, Merullo,
  Geiger, Lewis, McGrath, and Lubana}]{sarfati2026shape}
Rapha{\"e}l Sarfati, Eric Bigelow, Daniel Wurgaft, Siddharth Boppana, Jack
  Merullo, Atticus Geiger, Owen Lewis, Tom McGrath, and Ekdeep~Singh Lubana.
  2026.
\newblock \href {https://arxiv.org/abs/2602.02315} {The shape of beliefs:
  Geometry, dynamics, and interventions along representation manifolds of
  language models' posteriors}.
\newblock \emph{Preprint}, arXiv:2602.02315.

\bibitem[{Shai et~al.(2026)Shai, Amdahl-Culleton, Christensen, Bigelow, Rosas,
  Boyd, Alt, Ray, and Riechers}]{shai2026factored}
Adam Shai, Loren Amdahl-Culleton, Casper~L. Christensen, Henry~R. Bigelow,
  Fernando~E. Rosas, Alexander~B. Boyd, Eric~A. Alt, Kyle~J. Ray, and Paul~M.
  Riechers. 2026.
\newblock \href {https://arxiv.org/abs/2602.02385} {Transformers learn factored
  representations}.
\newblock \emph{Preprint}, arXiv:2602.02385.

\bibitem[{Shai et~al.(2024)Shai, Marzen, Teixeira, Oldenziel, and
  Riechers}]{shai2024belief}
Adam~S. Shai, Sarah~E. Marzen, Lucas Teixeira, Alexander~Gietelink Oldenziel,
  and Paul~M. Riechers. 2024.
\newblock \href {https://arxiv.org/abs/2405.15943} {Transformers represent
  belief state geometry in their residual stream}.
\newblock \emph{Preprint}, arXiv:2405.15943.

\end{thebibliography}

\end{document}